\documentclass{article}
\usepackage{scml2024}

\usepackage[utf8]{inputenc} 
\usepackage[T1]{fontenc}    
\usepackage{hyperref}       
\hypersetup{colorlinks,allcolors=blue}
\usepackage{url}            
\usepackage{booktabs}       
\usepackage{amsfonts}       
\usepackage{nicefrac}       
\usepackage{microtype}      
\usepackage{lipsum}         
\usepackage{graphicx}
\usepackage[numbers]{natbib}
\usepackage{doi}
\usepackage{wrapfig}
\usepackage{subfig}
\usepackage{xcolor}

\usepackage{float}

\title{Using Neural Implicit Flow to represent latent dynamics of canonical systems }

\date{}
\newif\ifuniqueAffiliation
\uniqueAffiliationtrue

\ifuniqueAffiliation 
\usepackage{authblk}

\author[1,3]{%
	Imran Nasim\thanks{\texttt{imran.nasim@ibm.com}, \texttt{i.nasim@surrey.ac.uk}}}%
\author[2]{%
	Joa\~o Lucas de Sousa Almeida\thanks{\texttt{joao.lucas.sousa.almeida@ibm.com}}}%
\affil[1]{IBM, UK}
\affil[2]{IBM Research Brazil}
\affil[3]{Department of Mathematics, University of Surrey, Guildford, GU2 7XH, Surrey, UK}

\begin{document}

\maketitle

\begin{abstract}

  The recently introduced class of architectures known as Neural Operators has emerged as highly versatile tools applicable to a wide range of tasks in the field of Scientific Machine Learning (SciML), including data representation and forecasting. 
In this study, we investigate the capabilities of Neural Implicit Flow (NIF), a recently developed mesh-agnostic neural operator, for representing the latent dynamics of canonical systems such as the Kuramoto-Sivashinsky (KS), forced Korteweg–de Vries (fKdV), and Sine-Gordon (SG) equations, as well as for extracting dynamically relevant information from them.
Finally we assess the applicability of NIF as a dimensionality reduction algorithm and conduct a comparative analysis with another widely recognized family of neural operators, known as Deep Operator Networks (DeepONets).
  
\end{abstract}

\section{Introduction}
\begin{figure}
  \centering
  \subfloat[NIF]{\includegraphics[scale=0.16]{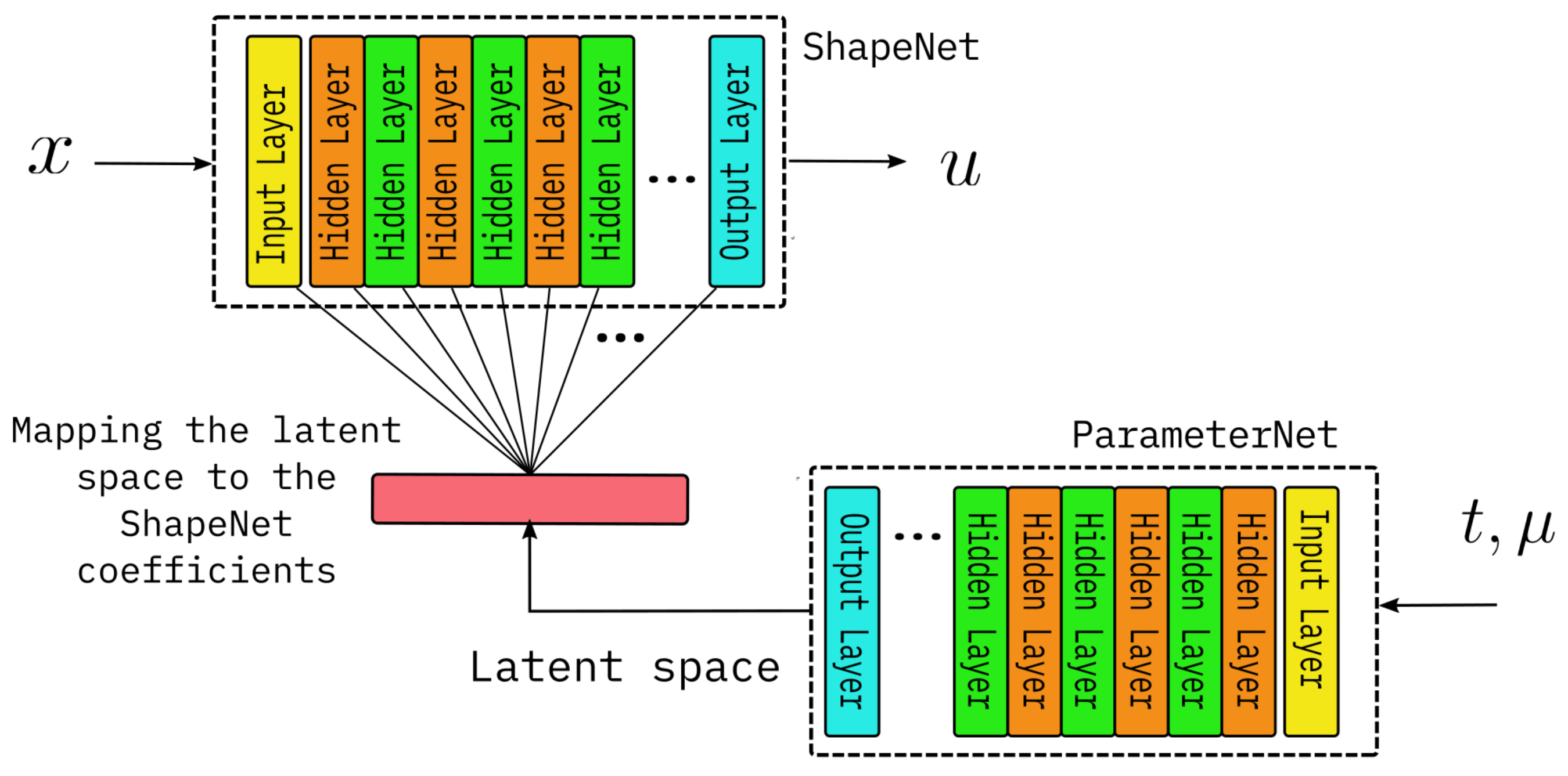}}
  \quad
  \vrule
  \quad
  \subfloat[DeepONet]{\includegraphics[scale=0.16]{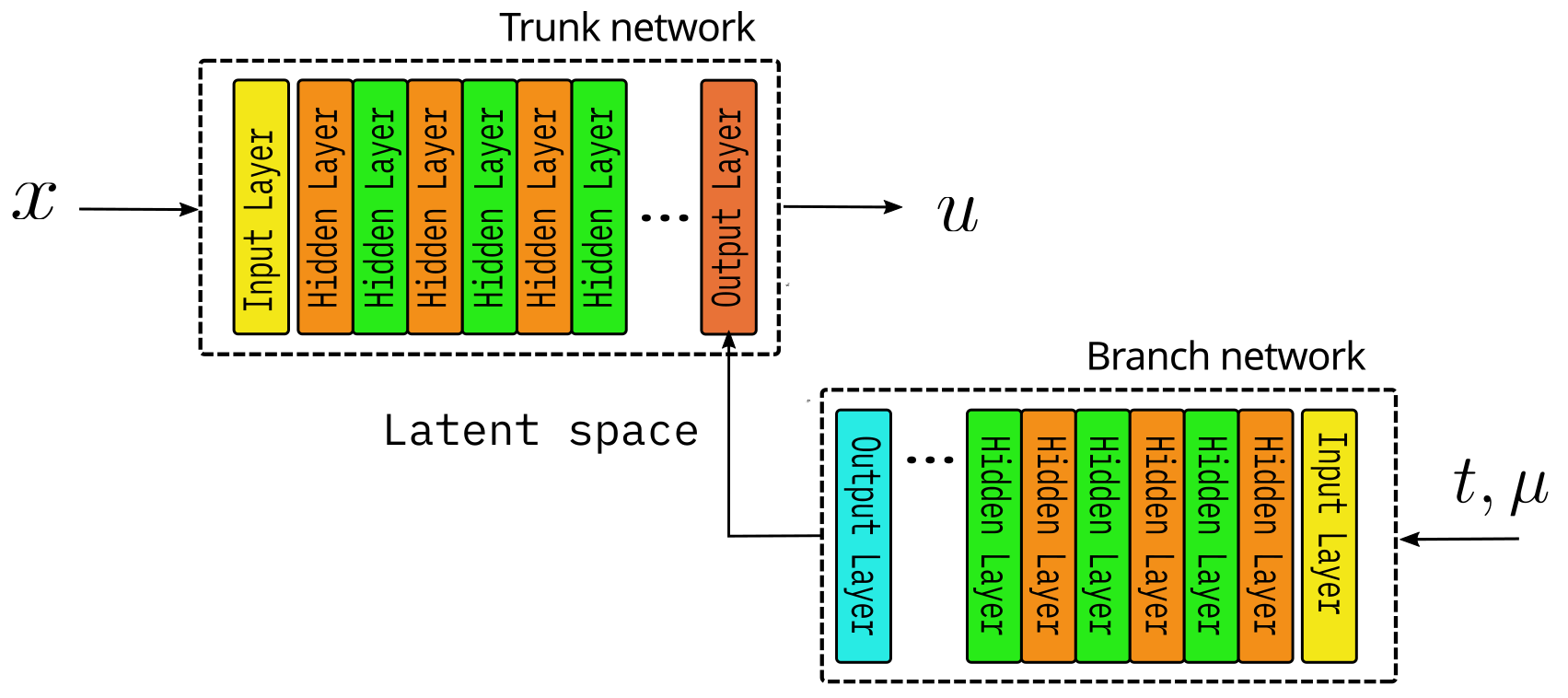}}
  \caption{NIF and DeepONet architectures.}
  \label{fig:nif_architecture}
\end{figure}

    Over the last few years the class of the so-called Neural Operators \cite{kovachki2023neural} have emerged as a promising tool for many fundamental tasks in scientific machine learning (SciML), as data representation \cite{shaowu}, time-series forecasting \cite{wang_perdikaris} and discovering of operators from data \cite{lu2021deeponet} both in data-driven and Physics-informed domains \cite{Wang2022,nasim2024_dynamically}. Neural Operators first appeared with the introduction of Deep Neural Operators (DeepONets) \cite{lu2021deeponet}, a new class of architectures designed to extend the capabilities of neural networks in order to  better perform tasks related to operator learning. 
A DeepONet is composed by two subnetworks, termed trunk and branch, and essentially emulates a linear expansion, in which the trunk learns a set of basis functions for a predetermined system of coordinates, while the branch discovers penalties for these functions as they relate to the forcing variables. Alternatively, it is possible to see the branch network as a hypernetwork aimed at evaluating the last layer for the trunk \cite{shaowu}.
Since DeepONets were first proposed, many derived and alternative approaches have been developed to address the operator learning problem. These approaches include novel ways to combine different neural network architectures within the DeepONet framework \cite{Oommen2022}, such as the Fourier Neural Operators (FNO) \cite{li2021fourier}, and more recently the Neural Implicit Flow (NIF) \cite{shaowu}, a hypernetwork which can be described as an extension of the original DeepONet concept (see Figure \ref{fig:nif_architecture}).
In order for a deep learning model to faithfully represent dynamics of an arbitrary dynamical system, it must be generalizable to systems that exhibit both quantitatively and qualitatively different dynamics which naturally present themselves in physical and engineering systems.
There exists a distinction in dynamical system models between integrable and non-integrable systems, both of which exhibit distinctly different dynamical behaviours \cite{ramani1989_integrability}.
Rather loosely defined, integrable systems have many conserved quantities confining motion to a lower-dimensional sub-manifold in the phase space. Conversely, non-integrable systems lack this characteristic giving rise to complex and chaotic motion \cite{fomenko2012_integrability}. In this work, we test the capabilities of NIF on both integrable and non-integrable systems namely the forced
Korteweg–de Vries (fKdV), Sine-Gordon (SG) and Kuramoto-Sivashinsky (KS) equations. We investigate whether the latent representation contains dynamically meaningful features and compare the results with DeepONets, a well known family of neural operators.

%
\section{Models and Experiments}\label{sec:mod_exp}
    \textbf{DeepONet.}
    \textcolor{black}{DeepONet \cite{lu2021deeponet} is a neural operator composed of two subnetworks, called trunk and branch, designed to universally approximate nonlinear operators. The branch network approximates a manifold representing the most fundamental system dynamics and the trunk projects the original spatiotemporal coordinates onto this latent manifold. 
In this work, we have employed DeepONets for a very specific task, that is to separate spatial and temporal components in order to emulate a proper orthogonal decomposition (POD). Such an approach makes it different from POD-DeepONet \cite{lu_pod_deeponet}, which uses POD as input for the trunk network, but it comes close to the so-called SVD-DeepONet \cite{venturi}), in which the neural network is employed to mimic a spectral expansion. The proposal is to use the branch network of the DeepONet to extract latent series from the system, which can be used for representing fundamental dynamics.\newline }\newline
   \textbf{NIF.}
   NIF is composed by two subnetworks, termed ParameterNet and ShapeNet, which closely resembles the branch and trunk components of DeepONets. However, differently from the branch network, ParameterNet evaluates not only the last layer but the entire set of weights and biases for ShapeNet enabling the model to mimic a decoder for each parametric input.
In order to avoid issues with the large number of ParameterNet outputs required to estimate all the coefficients for ShapeNet, a linear operation  (see Figure \ref{fig:nif_architecture}) is used to project the latent space onto the expected set of weights and biases \cite{shaowu}.
NIF is intended to work as a dimensionality reduction algorithm, since the ParameterNet creates a low-dimensional representation of the full-space dataset, which can be used for further modelling, as sparse reconstruction \cite{shaowu}, forecasting and even data generation. NIF can also serve as a representation model \cite{sitzmann} able of compressing information about large datasets into a relatively small set of parameters. 
All of the NIF models considered in this study were run for 5000 epochs with a learning rate of 5e-3, having 2 layers with 30 units in both the ShapeNet and ParameterNet. These hyperparameter values are within the ranges tested \cite{shaowu}. The DeepONet used to create the example in Figure \ref{fig:latent_spaces} is composed by two networks having three layers with 20 units and a sine activation. The DeepONet was trained for 5000 epochs with a learning rate of 1e-3.

   \textbf{Test cases and datasets.}
   The forced Korteweg–de Vries (fKdV) equation is an integrable non-linear PDE that is often used to model the weakly non-linear flow problem \cite{binder2005_fkdv,fkdv_2019}, which describes the existence of multiple possible solutions which is dependent on the type of disturbance. In our study,  we will consider the fKdV under the assumption of no disturbance. Under this assumption, the equation can be written as $6u_t + u_{xxx} + (9u - 6(F-1))u_x = 0$,
where $F$ is the depth-based Froude number. This model has been shown to exhibit both periodic travelling wave and soliton dynamics \cite{fkdv_2019,nasim2024_dynamically}. The second model we consider is the Kuramoto--Sivashinksy equation (KS hereafter).
The KS equation is a fourth-order non-integrable nonlinear PDE that is often used to model the evolution of surface waves and pattern formation for a number of physical systems \cite{kuramoto_1978}. The viscous form of the equation can be written as $ u_{t} + uu_{x} + u_{xx} + \nu u_{xxxx} = 0$,
where $\nu$ is a coefficient of viscosity.
The KS equation is highly complex, capturing the dynamics of spatio-temporal instabilities seen in a number of fluid flows. It also displays chaotic motion and due to its non-integrability gives rise to a rich array of solution types depending on the value of the viscosity parameter $\nu$. Pertinent to this study, the KS equation exhibits bursting wave dynamics for $\nu = \frac{16}{71}$ \citep{kirby1992_ks_bursting}. We also consider the Sine Gordon (SG) equation which gained popularity for the ability to exhibit soliton solutions \citep{hirota1972_SG} and can be written as $u_{tt} - u_{xx} + \sin{x} = 0.$
We simulate the fKdV and KS equations on a grid of 64 points on a periodic domain of $-\pi \leq x \leq \pi$ and set the Froude number $F=1.5$. We consider initial condition of the form $u(x,0) = {A}\cos\left({k}x + \phi\right)$ where we fix $A=0.5$, $k=1$ and $\phi=1$. We collect the dynamical data for all models from time $t=300$, defined as $T$, which is well after the initial transients have died out in order to capture the dynamics on the attractor.
All integrations for the KS and fKdV were performed using an explicit RK finite-difference scheme with a tolerance of $10^{-6}$ which was compared to a psuedo-spectral method to ensure accuracy. To solve the SG equation, we used an adaptive 5th order exponential time differencing method \citep{whalen2015_etd}.

\section{Results}
\begin{figure}[htb]
    \centering    
    \includegraphics[width=0.80\textwidth]{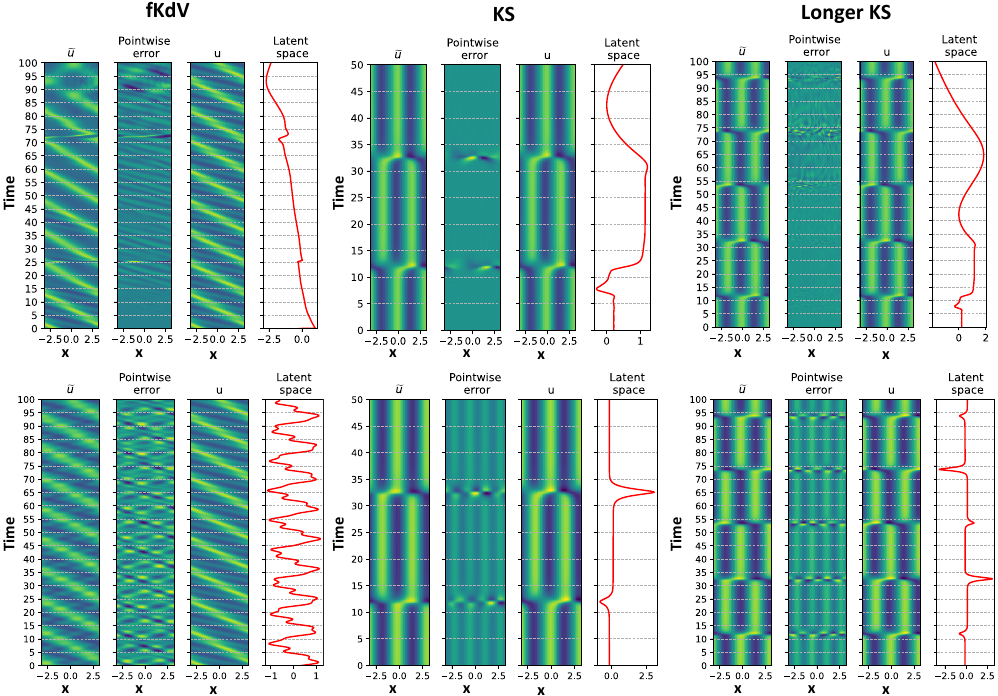}
    \caption{Latent variable profile with the pointwise error and predicted dynamics using NIF (upper panels) and DeepONet (lower panels).}
    \label{fig:latent_plot}
\end{figure}

   We first test whether NIF is able to faithfully predict the traveling wave data from the fKdV equation, the results for NIF and the comparative DeepONet model is presented in Figure \ref{fig:latent_plot}. 
We observe that the predictions from NIF are able to accurately predict the travelling wave dynamics for majority of the time domain yielding a smaller pointwise error compared to DeepONets. There appears to be deviation from the true data at later times, which is most evident for $T>90$, though the overall predictions are more accurate than the DeepONet model. Using the same numerical setup, we proceed to test NIF on the more complex bursting dynamics of the KS equation. We obtain the bursting dynamics data after the transients have died out in the same way as in the fKdV case which captures fast transitions between two saddle points connected by four heteroclinic connections \cite{kevrekidis1990_ks,nasim2024_dynamically}.  
From Figure \ref{fig:latent_plot} we see that NIF performs extremely well on predicting the bursting dynamics of the KS equation with essentially negligible error throughout the entire time domain\footnote{From the experiments we conducted we found that the Swish activation function provided the most accurate prediction of the bursting and travelling wave dynamics.}. We observe that the minor deviations present in the prediction occur at the times of rapid transition in the true data which is due to the aforementioned transition between saddle points and appear to get progressively stronger with time. Interestingly, we also observe this characteristic in the profiles obtained from the DeepONet model.
An important feature of any deep learning based surrogate is having a meaningful compact latent representation with the aim to capture the underlying dynamics that govern the system's behavior. To probe the nature of the latent representation in NIF we plot in Figure \ref{fig:latent_plot} the evolution of the latent variable as a function of time along with the reconstruction and pointwise error for all the models.
The middle panel shows the evolution of the latent variable for the case of the KS bursting dynamics. We clearly observe that transitions occur in the latent profile at the times where the phase transitions occur in the numerical solution suggesting that the latent variable is capturing these transitions that are occurring. Interestingly, when we probe the latent variable profile on the longer bursting dynamics data, right panel of Figure \ref{fig:latent_plot}, we notice that the strength of these transitions appear to decrease with time. In the case of the latent profile for the fKdV travelling wave data, it appears that the latent variable decreases steadily with time only having deviations from this behaviour at later times where the prediction error from NIF is larger. This profile is significantly different to that obtained by DeepONets what appears to capture a quasi-periodic behaviour which is more interpretable and explainable than the NIF representation.

\begin{figure}[!ht]
    \centering
    \subfloat{\includegraphics[scale=0.47]{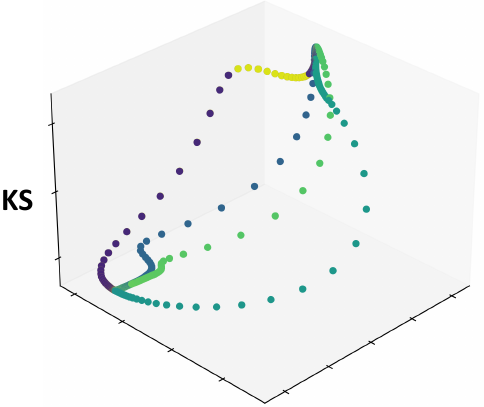}}
    \subfloat{\includegraphics[scale=0.28]{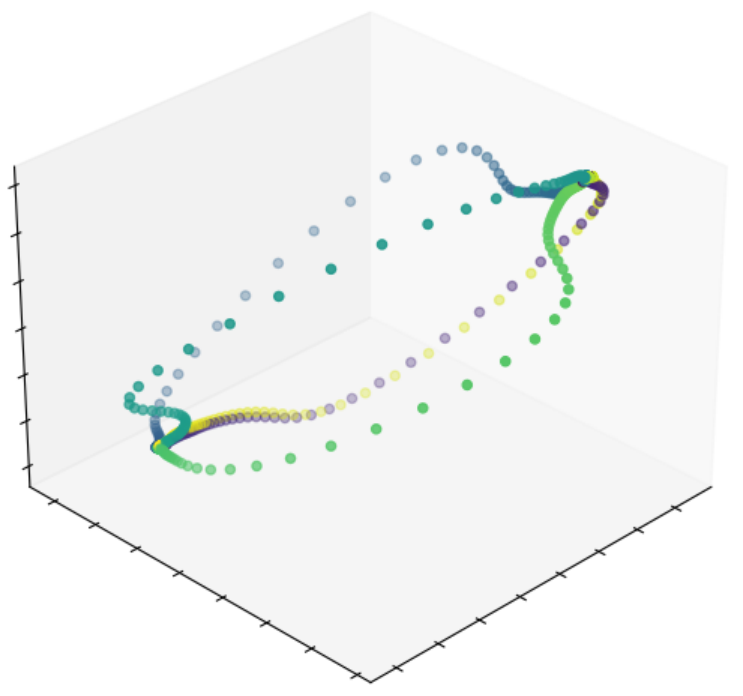}}
    \subfloat{\includegraphics[scale=0.28]{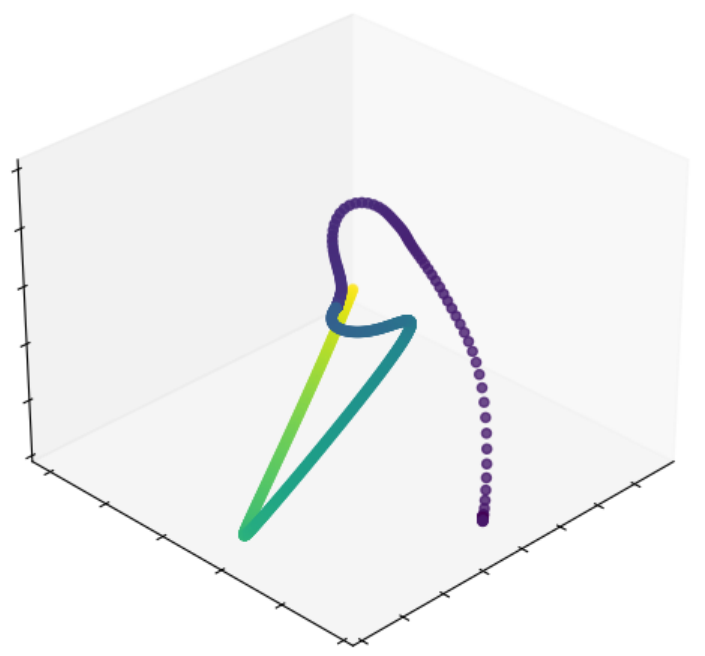}}\\
    \subfloat{\includegraphics[scale=0.66]{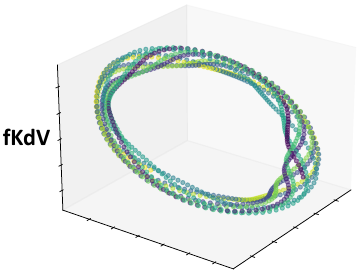}}
    \subfloat{\includegraphics[scale=0.28]{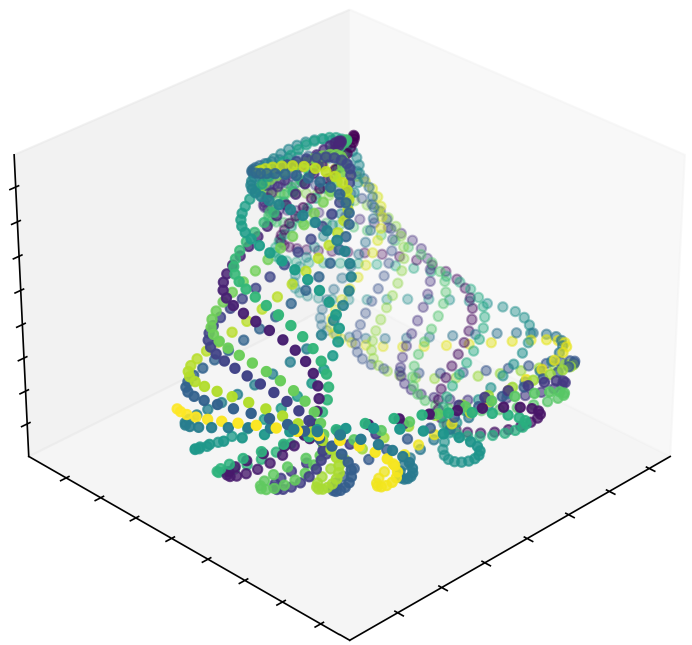}}
    \subfloat{\includegraphics[scale=0.28]{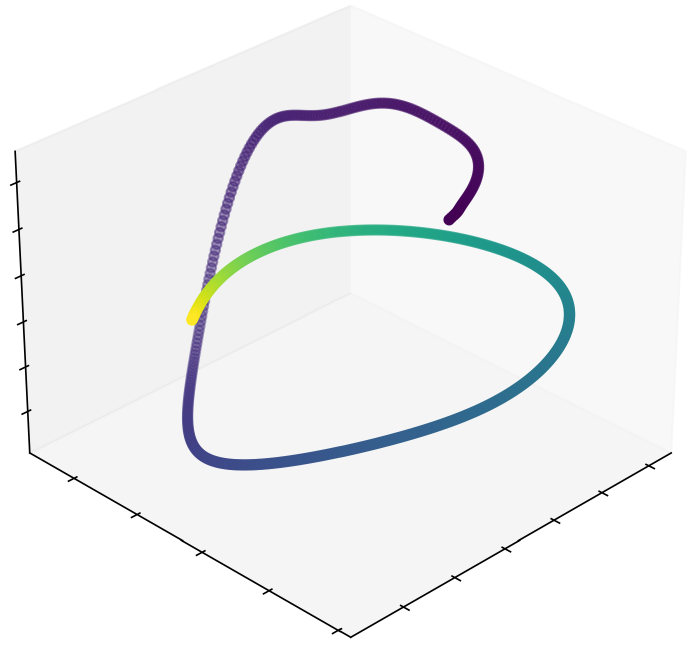}}
    \caption{The three-dimensional latent dynamics for KS (top) and fKdV (bottom) bursting dynamics data produced by i) Fourier projection (left panel) ii) DeepONet (middle panel) and iii) NIF (right panel).}
    \label{fig:latent_spaces}
\end{figure}

To further probe the nature of the captured latent representation, we proceed to run NIF with 3 latent variables and compare this representation with that obtained by DeepONets. To provide a reference `ground truth' latent representation for comparison we use the representations obtained by a Fourier projection of the full order data on the two dominant modes which highlights i) the skeletal attractor for the KS dynamics ii) quasi-periodic torus for the fKdV data \citep{nasim2024_dynamically} (see left panel of Figure \ref{fig:latent_spaces}). Interestingly in Figure \ref{fig:latent_spaces} we observe that the latent representation captured by DeepONets is in full agreement with the Fourier projection, both of which capture the transition between two saddle points via four heteroclinic connections \citep{kevrekidis1990_ks,nasim2024_dynamically}. We do not find the representation obtained by NIF to be dynamically interpretable. \textcolor{black}{Similarly in Figure \ref{fig:latent_spaces}, we find a good level of explainability from the DeepONet latent representation, capturing the quasi-periodicity and transition regions unlike the representation yielded from NIF which doesn't appear to capture this phase transition behaviour.
The results for the SG high frequency plane wave data is presented in Appendix \ref{appendix}. Again we find that DeepONets yields a more interpretable latent representation but has a substantially higher reconstruction error compared to NIF, an effect observed for all models.
A point worth noting is that there is a substantial effect of the latent space dimension for DeepONets, for example the reconstruction error decreases from $32 \%$ to $6 \%$ when this dimension is increased from $3$ to $6$. This may signify the ability for DeepONets to find the intrinsic manifold dimensionality, an effects that has been observed with autoencoders \citep{floryan2022_intrinsic}}. While we have evidence that the latent variables in NIF are sensitive to transitions in the raw data, overall we find the representation yielded by NIF in comparison to DeepONets to be significantly less interpretable. We find these results very surprising given the relative reconstruction errors: for the KS bursting dynamics data, $12.3 \%$ for DeepONet and $1.68 \%$ for NIF. For the fKdV data, NIF achieves an error of $2.0\%$ whereas DeepONets yields $5.0\%$. Additionally for the SG data, NIF achieves an error of $2.68\%$ whereas DeepONets yields $7.48\%$.
A point worth noting here is that unlike DeepONets, NIF doesn't fully separate variables between its two sub-networks, with ShapeNet coefficients still relying on ParameterNet inputs. Although it hinders the analogy with a linear decomposition, as Fourier (see Figure \ref{fig:latent_spaces}), it can allow the model to discover the most proper modes to represent each snapshot, which endows it with a greater capacity to represent complex datasets. 

\section{Impact statement}
   In this study we have explored the capabilities of NIF for representing qualitatively distinct dynamics for both integrable and non-integrable PDE systems. We find the reconstructed dynamics predicted by NIF to be accurate and faithful to the dynamics present in the true data. This suggests that NIF can be used to represent datasets related to dynamical systems in a low-dimensional way, thus enabling us to extract important information from this latent dynamics.
While our results support the idea that the latent representation in NIF is sensitive to the dynamical characteristics present in the true data, its interpretability is not as intuitively clear when compared to representations yielded by a state of the art DeepONet architecture. We find this result surprising given the fact that NIF yields a significantly smaller reconstruction error on all datasets. A further investigation into interpretability by analyzing the NIF architecture and regularization methods will be investigated in a future study.

\bibliographystyle{plain}
\bibliography{bibliography}

\begin{thebibliography}{10}

\bibitem{fkdv_2019}
Benjamin~J. Binder.
\newblock Steady two-dimensional free-surface flow past disturbances in an open channel: Solutions of the korteweg–de vries equation and analysis of the weakly nonlinear phase space.
\newblock {\em Fluids}, 4(1), 2019.

\bibitem{binder2005_fkdv}
BJ~Binder and J-M Vanden-Broeck.
\newblock Free surface flows past surfboards and sluice gates.
\newblock {\em European Journal of Applied Mathematics}, 16(5):601--619, 2005.

\bibitem{floryan2022_intrinsic}
Daniel Floryan and Michael~D Graham.
\newblock Data-driven discovery of intrinsic dynamics.
\newblock {\em Nature Machine Intelligence}, 4(12):1113--1120, 2022.

\bibitem{fomenko2012_integrability}
Anatolij~T Fomenko.
\newblock {\em Integrability and nonintegrability in geometry and mechanics}, volume~31.
\newblock Springer Science \& Business Media, 2012.

\bibitem{hirota1972_SG}
Ryogo Hirota.
\newblock Exact solution of the sine-gordon equation for multiple collisions of solitons.
\newblock {\em Journal of the Physical Society of Japan}, 33(5):1459--1463, 1972.

\bibitem{kevrekidis1990_ks}
Ioannis~G Kevrekidis, Basil Nicolaenko, and James~C Scovel.
\newblock Back in the saddle again: a computer assisted study of the kuramoto--sivashinsky equation.
\newblock {\em SIAM Journal on Applied Mathematics}, 50(3):760--790, 1990.

\bibitem{kirby1992_ks_bursting}
Michael Kirby and Dieter Armbruster.
\newblock Reconstructing phase space from pde simulations.
\newblock {\em Zeitschrift f{\"u}r angewandte Mathematik und Physik ZAMP}, 43:999--1022, 1992.

\bibitem{kovachki2023neural}
Nikola Kovachki, Zongyi Li, Burigede Liu, Kamyar Azizzadenesheli, Kaushik Bhattacharya, Andrew Stuart, and Anima Anandkumar.
\newblock Neural operator: Learning maps between function spaces, 2023.

\bibitem{kuramoto_1978}
Yoshiki Kuramoto.
\newblock {Diffusion-Induced Chaos in Reaction Systems}.
\newblock {\em Progress of Theoretical Physics Supplement}, 64:346--367, 02 1978.

\bibitem{li2021fourier}
Zongyi Li, Nikola~Borislavov Kovachki, Kamyar Azizzadenesheli, Burigede liu, Kaushik Bhattacharya, Andrew Stuart, and Anima Anandkumar.
\newblock Fourier neural operator for parametric partial differential equations.
\newblock In {\em International Conference on Learning Representations}, 2021.

\bibitem{lu2021deeponet}
Lu~Lu, Pengzhan Jin, Guofei Pang, Zhongqiang Zhang, and George~Em Karniadakis.
\newblock Learning nonlinear operators via deeponet based on the universal approximation theorem of operators.
\newblock {\em Nature Machine Intelligence}, 3:218--229, 2021.
\newblock in press.

\bibitem{lu_pod_deeponet}
Lu~Lu, Xuhui Meng, Shengze Cai, Zhiping Mao, Somdatta Goswami, Zhongqiang Zhang, and George~Em Karniadakis.
\newblock A comprehensive and fair comparison of two neural operators (with practical extensions) based on fair data.
\newblock {\em Computer Methods in Applied Mechanics and Engineering}, 2021.

\bibitem{nasim2024_dynamically}
Imran Nasim and Michael~E Henderson.
\newblock Dynamically meaningful latent representations of dynamical systems.
\newblock {\em Mathematics}, 12(3):476, 2024.

\bibitem{Oommen2022}
Vivek Oommen, Khemraj Shukla, Somdatta Goswami, R{\'e}mi Dingreville, and George~Em Karniadakis.
\newblock Learning two-phase microstructure evolution using neural operators and autoencoder architectures.
\newblock {\em npj Computational Materials}, 8(1):190, Sep 2022.

\bibitem{shaowu}
Shaowu Pan, Steven~L. Brunton, and J.~Nathan Kutz.
\newblock Neural implicit flow: a mesh-agnostic dimensionality reduction paradigm of spatio-temporal data.
\newblock {\em Journal of Machine Learning Research}, 24(41):1--60, 2023.

\bibitem{ramani1989_integrability}
Alfred Ramani, Basil Grammaticos, and Tassos Bountis.
\newblock The painlev{\'e} property and singularity analysis of integrable and non-integrable systems.
\newblock {\em Physics Reports}, 180(3):159--245, 1989.

\bibitem{sitzmann}
Vincent Sitzmann, Julien Martel, Alexander Bergman, David Lindell, and Gordon Wetzstein.
\newblock Implicit neural representations with periodic activation functions.
\newblock In H.~Larochelle, M.~Ranzato, R.~Hadsell, M.F. Balcan, and H.~Lin, editors, {\em Advances in Neural Information Processing Systems}, volume~33, pages 7462--7473. Curran Associates, Inc., 2020.

\bibitem{venturi}
Simone Venturi and Tiernan Casey.
\newblock Svd perspectives for augmenting deeponet flexibility and interpretability.
\newblock {\em Computer Methods in Applied Mechanics and Engineering}, 403:115718, 2023.

\bibitem{wang_perdikaris}
Sifan Wang and Paris Perdikaris.
\newblock Long-time integration of parametric evolution equations with physics-informed deeponets.
\newblock {\em Journal of Computational Physics}, 475:111855, 2023.

\bibitem{Wang2022}
Sifan Wang, Hanwen Wang, and Paris Perdikaris.
\newblock Improved architectures and training algorithms for deep operator networks.
\newblock {\em Journal of Scientific Computing}, 92(2):35, Jun 2022.

\bibitem{whalen2015_etd}
Patrick Whalen, Moysey Brio, and Jerome~V Moloney.
\newblock Exponential time-differencing with embedded runge--kutta adaptive step control.
\newblock {\em Journal of Computational Physics}, 280:579--601, 2015.

\end{thebibliography}

\appendix
\section{Results for Sine-Gordon Equations}\label{appendix}
The neural networks used for Sine-Gordon have the same configurations described in Section \ref{sec:mod_exp}. For DeepONet, the reconstruction error using three latent variables is $7.48 \%$ and for NIF is $2.68\%$. 
\begin{figure}[H] 
    \centering
    \subfloat{\includegraphics[scale=0.47]{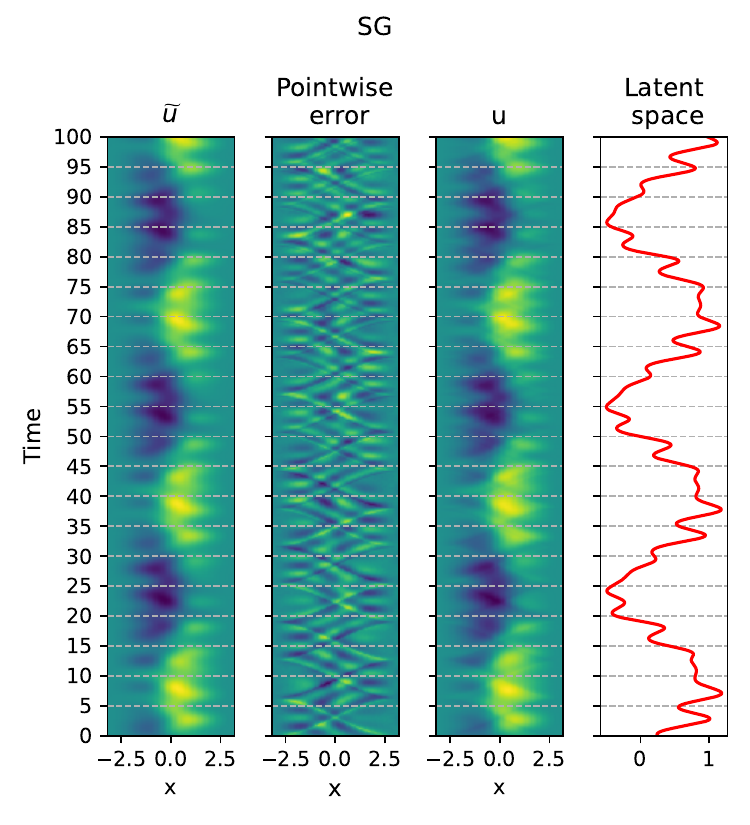}}
    \subfloat{\includegraphics[scale=0.47]{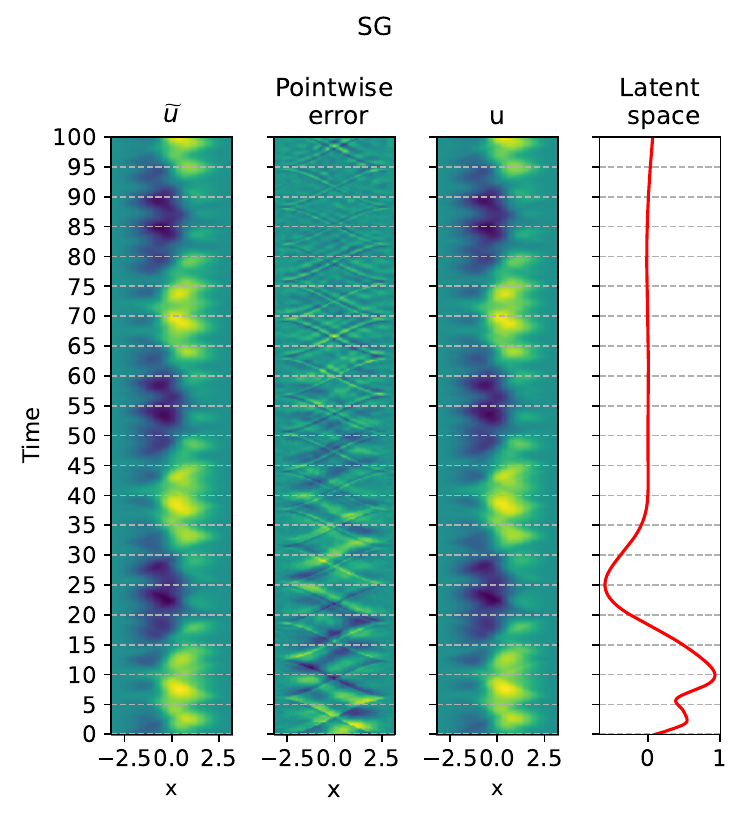}}
    \caption{Latent variable profile with the pointwise error and predicted dynamics using i) DeepONet (left panel) and ii) NIF (right panel).}
\end{figure}
\begin{figure}[H]
    \centering
    \subfloat{\includegraphics[scale=0.67]{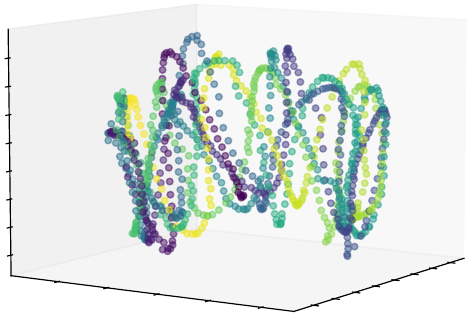}}
    \subfloat{\includegraphics[scale=0.47]{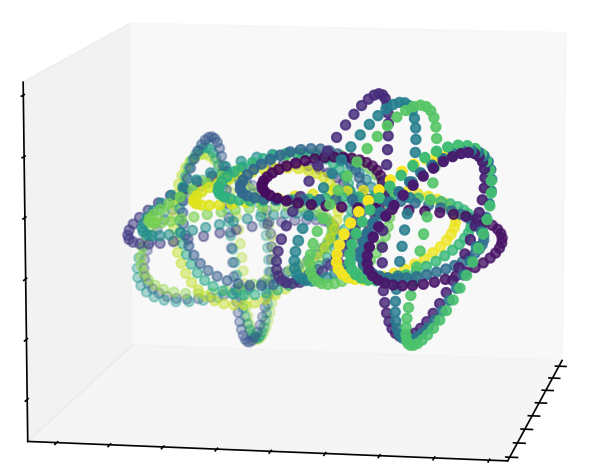}}
    \subfloat{\includegraphics[scale=0.47]{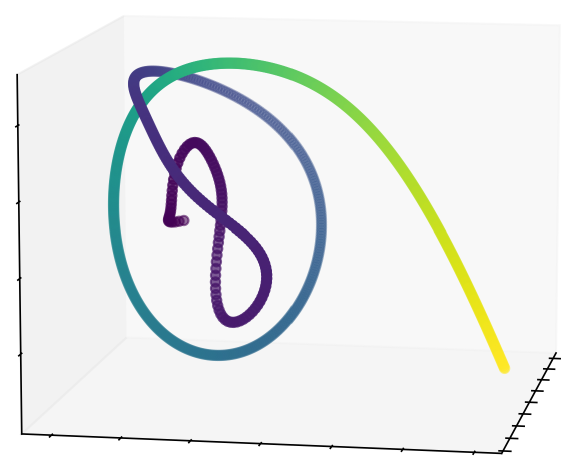}}
    \caption{The three-dimensional latent dynamics for Sine-Gordon bursting dynamics data produced by i) Fourier projection (left panel) and ii) DeepONet (middle panel) and iii) NIF (right panel).}
\end{figure}

\end{document}